\documentclass[runningheads]{llncs}
\usepackage[T1]{fontenc}
\usepackage{graphicx}
\usepackage{booktabs}
\usepackage[misc]{ifsym}

\usepackage{amsmath}
\usepackage{multirow}
\usepackage{hyperref}

\newcommand{\best}[1]{\textbf{#1}}
\newcommand{\second}[1]{\underline{#1}}

\begin{document}

\title{Simulating Biases for Interpretable Fairness in Offline and Online Classifiers\thanks{This work was partially funded by projects AISym4Med (101095387) supported by Horizon Europe Cluster 1: Health, ConnectedHealth (n.º 46858), supported by Competitiveness and Internationalisation Operational Programme (POCI) and Lisbon Regional Operational Programme (LISBOA 2020), under the PORTUGAL 2020 Partnership Agreement, through the European Regional Development Fund(ERDF) and Agenda “Center for Responsible AI”, nr. C645008882-00000055, investment project nr. 62, financed by the Recovery and Resilience Plan (PRR) and by European Union - NextGeneration EU, and also by FCT plurianual funding for 2020-2023 of LIACC (UIDB/00027/2020 UIDP/00027/2020). Funded by the European Union – NextGenerationEU. Views and opinions expressed are however those of the author(s) only and do not necessarily reflect those of the European Union or the European Commission. Neither the European Union nor the European Commission can be held responsible for them.}}
\titlerunning{Simulating Biases for Interpretable Fairness in Classifiers}

\author{Ricardo In\'acio\inst{1,2} \and Zafeiris Kokkinogenis\inst{1,2} \and Vitor Cerqueira\inst{1,2} \and \\Carlos Soares\inst{1,2}}

\institute{Faculdade de Engenharia da Universidade do Porto, Portugal\\ \email{\{rcinacio, zafeiris, vcerqueira, csoares\}@fe.up.pt}\\ 
\and Laboratory for Artificial Intelligence and Computer Science (LIACC), Portugal}
%
\maketitle              
\setcounter{footnote}{0} 
\begin{abstract}

Predictive models often reinforce biases which were originally embedded in their training data, through skewed decisions. In such cases, mitigation methods are critical to ensure that, regardless of the prevailing disparities, model outcomes are adjusted to be fair.
To assess this, datasets could be systematically generated with specific biases, to train machine learning classifiers. Then, predictive outcomes could
aid in the understanding of this bias embedding process.
Hence, an agent-based model (ABM), depicting a loan application process that represents various systemic biases across two demographic groups, was developed to produce synthetic datasets. Then, by applying classifiers trained on them to predict loan outcomes, we can assess how biased data leads to unfairness. This highlights a main contribution of this work: a framework for synthetic dataset generation with controllable bias injection. We also contribute with a novel explainability technique, which shows how mitigations affect the way classifiers leverage data features, via second-order Shapley values.
In experiments, both offline and online learning approaches are employed. Mitigations are applied at different stages of the modelling pipeline, such as during pre-processing and in-processing. 
        
\end{abstract}

\section{Introduction}\label{sec:intro}

Machine learning models are crucial in several domains to ensure a proper data-driven decision-making process. Nonetheless, several data-related factors can severely degrade predictive performance. In classification tasks, if the target distribution is skewed, predictions regarding the minority might not be as reliable as the ones for the majority~\cite{haixiang2017learning}. Also, if certain biases are concealed in the data, not only will these be reflected in predictions, but also further amplified~\cite{wang2021directional}. These might surface when the group often denoted as protected, is characterised by sensitive features (e.g., ethnicity) that expose them to unfair treatment~\cite{goethals2024beyond}. 

To further investigate this, 
synthetically biased datasets could be generated, to more clearly explain how it is embedded and reflected in classifiers. To that end, an Agent-Based Model (ABM) was developed to depict a loan application process, showcasing systemic biases across two demographic groups. Both partake in the same activities, while having disparate feature values that generally favour one over the other. The corresponding data is collected to develop two kinds of classifiers: offline (i.e., batch-based) and online (i.e., mini-batch-based). 

The goal is to showcase how biased data exacerbates disparities across groups. Metrics commonly used in machine learning tasks, although still relevant, might not be always reliable for biased scenarios, as these conceal group-wise disparities. Thus, metrics such as Equal Opportunity Difference (EOD)~\cite{huang2022evaluation}, are also employed. Mitigation methods of varying complexity, such as class-based reweighing, and constrained Exponentiated Gradient estimators, are applied. Then, the impact of mitigation methods on the way classifiers leverage data features, can be explained through Shapley values~\cite{lundberg2017unified,muschalik2024shapiq}. The main contributions of this paper are twofold. A novel ABM framework for the generation of synthetic datasets with controllable bias injection, to support model fairness auditing. And an explainability approach for the impact of different mitigation methods on classifiers, based on second-order Shapley values of feature interactions.

    \section{Literature Review}
    \label{sec:literature}

    This section outlines the background of this work, on agent-based models, offline and online algorithms, and how the biases embedded in datasets might affect predictive models. We then explore bias mitigation and explainability methods.

    \subsection{Agent-based Models}

    Agent-based models (ABM) systematically represent interactions in environments consisting of multiple entities, allowing for an analysis of the structures and dynamics it simulates~\cite{doi:10.1073/pnas.082080899}.
    The fundamental element is the agent, defined by a set of attributes whose values determine its individual characteristics. At any given point, the set of specific attribute values of an entity reflects its momentary state~\cite{de2014agent}. Agents also embody behaviours, that establish how they are able to engage with other agents, and the environment.

    The origins of ABM research is often attributed to John von Neumann and Stanislaw Ulam, in \textit{cellular automata} and self-replicating systems, where complex patterns would arise from simpler interactions. For social sciences, such methods are natural abstractions to explain emergent social phenomena. For example, in the foundational work by Schelling, agents in a shared-space model, that reveal an affinity towards others of the same group, eventually lead to segregated neighbourhoods~\cite{schelling1971}. Also, the Sugarscape model by Epstein and Axtell, which built upon earlier work, highlights how even if starting conditions are equal resource-wise, entities boasting enhanced attributes, such as vision and lower metabolism accumulate more resources~\cite{epstein1996growing}.

    \subsection{Batch Offline and Incremental Online Learning}
    \label{sec:online-batch}

    In traditional supervised machine learning, a fixed batch of data is collected, to then be modelled alongside the labels that characterise each instance. However, recent advancements in data-related technologies, which increased both its availability and granularity, highlighted their limitations to adapt to conceptual or distribution changes~\cite{klinkenberg2004learning}. This might make them improper for streaming contexts, given that the characteristics ought to naturally evolve~\cite{perez2018review}.

    To address these issues, algorithms that incrementally learn the patterns in data as it becomes available, updating model parameters in an incremental manner~\cite{hoi2021online} were introduced. While the term online learning refers to the overall methodology behind such algorithms, a distinction can be made based on the rate of the carried updates. Fully online algorithms update the model parameters on the arrival of each single instance~\cite{hoi2021online}. Semi-online algorithms are those that update after processing a mini-batch of data~\cite{zhang2020deterministic}, being these the preferred choice for learning deep networks, given higher training stability.

    \subsection{Bias in Machine Learning}
    \label{sec:bias-ml}

    When machine learning models are trained on data that conceals biases, such systems might reinforce and even amplify them~\cite{ntoutsi2020bias}, mainly through sensitive features, indicative of groups prone to prejudices (e.g., ethnicity). Yet, even when those are not directly present, other variables that might be strongly correlated may serve as proxy attributes for discrimination~\cite{wang2023overwriting} (e.g., address). Hence, the group characterised by them is denoted as protected, while those that are not are denoted as privileged~\cite{goethals2024beyond}. Also, if the data is not representative of all groups, it may lead to poor predictive performance towards minorities, which further aggravates the previous prejudices.
    
    Moreover, in the review conducted by Mehrabi et al.,~\cite{mehrabi2021survey}, the notions of \textit{explainable} and \textit{unexplainable} discrimination are laid out. The first refers to cases where discrepancies arise from well-understood circumstances, grounded on some rule or accepted phenomena, while the latter relates to prejudices grounded on illegal or discriminatory practices that promptly harm specific groups. These can be either \textit{direct}, where specific attributes explicitly lead to unfair treatment, or \textit{indirect}, which may arise from proxies, as previously mentioned.

    \subsection{Methods for Bias Mitigation and Explanation}
    \label{sec:metrics-mitigation}
    
    Mitigation approaches can be categorized by the step of the modelling pipeline that they can be applied. Pre-processing methods, provide a model-agnostic approach to balance datasets. A common strategy is data resampling, either by augmenting the minority class, reducing the majority, or both in tandem. Another method, entails assigning different weights to instances based on class membership, which benefits from not compromising relevant information~\cite{ntoutsi2020bias}.
    
    In-processing approaches, affect the underlying learning mechanisms during training. One example is Exponentiated Gradient reduction~\cite{weerts2023fairlearn}, which subjects a classifier to a given constraint, such as Demographic Parity, forcing it to carry out fair perditions irrespective of sensitive attributes or class belonging~\cite{agarwal2018reductions}. 
    
    Moreover, even after a successful mitigation (e.g., improved scores on fairness metrics), it may still be difficult to understand the real impact on the underlying model. Hence, explainability methods, such as Shapley values~\cite{lundberg2017unified}, can be employed. This approach explains how data features contribute individually, or more recently, in a pairwise manner~\cite{muschalik2024shapiq}, to reach specific outcomes, showcasing how classifiers leverage them in order to carry out predictions.
    
    \section{Conceptual Model}
    \label{sec:methodology}

    In this section, we lay out the core elements that constitute the ABM employed throughout this work, including the entities that participate in simulations, boasting shared and class-specific attributes. The model variables that characterize the biased scenarios used in the experiments are also shown.

    \subsection{Objectives}
	\label{sec:goals}

    The main goal is to model socioeconomic disparities, simulating a loan application process with sustained bias, and occasional monetary transactions across peers. We aim to understand and evaluate bias propagation, which is then reflected in the collected data. Also, we seek to illustrate how training classifiers on this data results in skewed assessments. We expect that although global metrics may showcase strong overall performance, group-level evaluation ought to show signs of inequality. We implement robust classifiers to predict the result of loan applications, aiming to show that although training is properly conducted, predictions will still favour the privileged group, solely based on biased data.

    \subsection{Simulated Entities}
	\label{sec:entities}

    Simulations are populated by two demographics: the group of privileged entities (A), and the group of protected entities (B). In each, members are endowed with a wide range of attributes, establishing the initial levels of structural, socio-economic, and historical biases, defining both explicit and implicit disparities.

    \subsubsection{Attributes.}
    
    When an entity is created, its group is randomly assigned based on the probability of representation (\texttt{rep}). Then, individual attributes are set based on group-specific ranges and probabilities, generally more favourable for entities belonging to group A. For example, \texttt{wealth} is uniformly sampled from a range of [50, 89] for group A, while for group B the range is [30, 59]. Regarding probabilities, the values for P(\texttt{has\_job}), P(\texttt{has\_car}), and P(\texttt{has\_house}) are respectively 0.9, 0.8, 0.6 for group A, and 0.7, 0.5, 0.3 for group B. Besides these, second-order effects are introduced, by combining the values of other attributes. Examples are \texttt{trust}, computed using both the \texttt{education} level and \texttt{wealth}, or financial literacy (\texttt{fin\_lit}), computed via \texttt{education} level, \texttt{wealth}, and \texttt{trust}.

    After the process, a loan qualification score is calculated for each applicant, and if a predefined threshold is crossed, the \texttt{qualified}, and consequently \texttt{loan\_approved} labels are set as true, moving the entity to the \texttt{processed} state. To make the collected data less artificial and trivial to predict, we simulate measurement errors by flipping the assigned label with a probability of 5\%.

    \subsubsection{Peer Network.}

    Entities might also join a peer network, represented as an undirected graph, if their \texttt{trust} is high enough. This network will ultimately be mostly composed of group A entities, based on generally higher trust values. Each participant is connected to at most three other entities, and the value of their trust attribute is then updated in the direction of the average trust of its peers (i.e., if trust is high among peers, one should also be more trustful). This means that clusters composed mostly of group A entities will form "high-trust societies", reinforcing group-based disparities through the network.

    There is also a chance in each time step of a transaction across entities. Given that the network will mostly embrace wealthier individuals, and since protected entities (B) have not only lower chances of entering it (i.e., based on structurally lower trust), but also systemically lower wealth, these will inadvertently benefit less from the conducted exchanges.
    
    \subsection{Variables of the System}
	\label{sec:variables}
    
    These variables specify the properties of the model itself, which in turn affect or are affected by every entity in the simulation. Through them, we specify the distinct scenarios we wish to simulate, in order to explicitly or implicitly introduce biases into the system.
    
    One example is the "\texttt{lbl}" feature, which specifies both the boost $\beta$ applied to the score of class A applicants: $\text{score} = \text{score} \times (1 + \beta)$, and the penalty applied to the score of class B applicants: $\text{score} = \text{score} \times (1 - \beta)$, directly skewing the final decision process, introducing historical or prejudice biases~\cite{mehrabi2021survey}.
    
    Another frequent issue in machine learning tasks is target imbalance~\cite{hoens2013imbalanced}, which might lead to representation or sampling biases~\cite{mehrabi2021survey}. If one of the groups is underrepresented, estimations for the related instances will ultimately be poor. Hence, the "\texttt{rep}" attribute allows us to define the probability for generating group A entities ($\alpha$), and consequently, the probability for group B ($1-\alpha$).

	\section{Experimental Design}
	\label{sec:operations}
    
    In this section we detail the simulation scenarios carried out in the conducted experiments, addressing both class-inherent biases conditioned on socio-economical factors, and the impact of an imbalanced group representation.
         
    \subsection{Operation Policies}
     
    We simulate the same scenarios in both modalities, differing only in how data is made available to the respective classifier: in a single batch for offline, and incrementally (i.e., mini-batches of 100 instances) for online. The ABM is executed for 10000 time steps in each, using different combinations of representation (\texttt{rep}) and prejudice (\texttt{lbl}) biases, explicitly including them. Specifically, the following values were used for \texttt{lbl}: [0.0, 0.4, 0.5, 0.6], and for \texttt{rep}: [0.5, 0.6, 0.7, 0.8], which led to 16 distinct scenarios. Notably, the pair (\texttt{lbl} = 0.0, \texttt{rep} = 0.5) represents a case where no explicit biases are applied (i.e., representation is balanced and no label boost or penalty is applied). Each of the combinations was systematically paired with four different mitigation strategies: (i) no mitigation (baseline), pre-processing methods, such as (ii) automatic reweighing of instances or (iii) manual reweighing using group-specific weights, and an in-processing technique, through (iv) an exponentiated gradient reduction model with fairness constraints, via either demographic parity or equalized odds.

    \subsection{Evaluation Design}
    \label{sec:evaluation-design}
    
    Standard metrics such as accuracy, precision, recall, log loss, and ROC AUC were adopted, to quantify overall classifier performance. Yet, these represent the global performance of a method when in aggregate, which may not be reliable in skewed contexts, as results ought to be mostly conditioned on estimations regarding the majority, concealing disparities. Some enable class-wise performance estimations, allowing for a more detailed assessment of discrimination performance across groups. Even so, such metrics might not accurately represent how fair the model is, since these do not take into consideration sensitive attributes.
    
    Hence, metrics that put emphasis on how sensitive attributes condition predictions, were also employed: Equal Opportunity Difference (EOD), based on the notion of equality of opportunity~\cite{hardt2016equality}, which quantifies the difference in true positive rates (TPR) across groups; and Statistical Parity Difference (SPD)~\cite{savani2020intra}, which measures the difference in the likelihood of a positive outcome across groups. In both, a value of 0 indicates perfect adherence to the fairness criteria, and a deviation indicates the presence of bias. The sign of the differences further indicates the direction of biases: a negative value denotes a favour towards the privileged group, and a positive value a favouring of the protected one. Furthermore, approval rate, which measures the probability of entities from a group being classified positively (i.e., approved loan), is used. 
    
    We hide from the classifiers the features more tightly related with socio-economic class and the outcomes of the loan process, such as \texttt{wealth}, and \texttt{credit\_score}, to avoid leakage. Moreover, \texttt{education} and \texttt{trust} are also  hidden, as these are not only used in the computation of other second-order features, but the latter is also the main component of the peer network.
    
    \subsection{Simulation Tools}
	\label{sec:tools}

    We design the full simulation engine using Python, in which the participating agents are represented in a class, and the peer interaction network is implemented through the \texttt{networkx}~\cite{hagberg2008exploring} library. Regarding model training, a robust pipeline was implemented in both modalities, considering the specificities from each to ensure that the process is as analogous as possible, in baseline settings.
    
    \subsubsection{Offline (Static) Pipeline.}
    
    The \texttt{XGBClassifier}~\cite{chen2016xgboost} implementation from the \texttt{xgboost} package was employed. We scale the data using the \texttt{StandardScaler} method from the \texttt{scikit-learn} library to ease modelling. Then, the parameters of the classifier are tuned using \texttt{RandomizedSearchCV}, also from \texttt{scikit-learn}, using only 3 iterations to reduce the computational time of the experiments, which ought to be sufficient given learning task difficulty. 
    Finally, Isotonic Regression, via the \texttt{CalibratedClassifierCV} method from the same library, is applied to calibrate the probabilities of the predictions, aligning them to the real class likelihoods.

    \subsubsection{Online (Dynamic) Pipeline.}

    An incremental Hoeffding tree from the \texttt{river} library~\cite{montiel2021river} (\texttt{HoeffdingTreeClassifier}) was used, and its hyperparameters were tuned by sampling ten candidate configurations via \texttt{ParameterSampler} (from \texttt{scikit-learn}), also across 3 initial tuning runs with fixed seed, ensuring identical simulations. To apply analogous scaling to the features, the online \texttt{StandardScaler} from \texttt{river.preprocessing} was employed. The classifier processes each observation incrementally, updating its parameters on the most recent mini-batch of data, every time the "update interval" is reached. To obtain well‐calibrated probabilities, raw scores and true labels are buffered until the \texttt{calibrate\_interval} is reached, to then also fit an Isotonic Regression model.

    \subsection{Bias Mitigation Methods}
    \label{sec:bias-mitigation}

    Pre-processing mitigation is applied in the form of reweighing, both automatically, via the \texttt{aif360}~\cite{aif360-oct-2018} library, and manually, by specifying group-specific weights. However, the \texttt{reweighing} pre-processor from \texttt{aif360} cannot be reliably employed in the online modality, expecting global counts of group frequency to compute weights~\cite{aif360-oct-2018}. To that end, we introduce a stabilized online reweigher, which approximates the same weighting scheme of the former, introduced by Kamiran \& Calders~\cite{kamiran2012data}, adapted to use exponentially weighted moving averages (EMA). This fosters a simple adaptation to drift, by focusing on the most recent data. It updates the total, target, group, and joint counts on each instance, to obtain the respective weights $w(g,y)$ (c.f. Equation~\ref{eq:weighting}). 

   {\small
    \begin{equation}
        w(g, y) = \mathrm{clip}\left(
        \frac{P_{\mathrm{EMA}}(y) \cdot P_{\mathrm{EMA}}(g)}{P_{\mathrm{EMA}}(g, y)},
        \; w_{\min}, \; w_{\max}
        \right)
        \label{eq:weighting}
    \end{equation}
    }

    In-processing mitigation is done through the \texttt{fairlearn}~\cite{weerts2023fairlearn} library, using  the \texttt{ExponentiatedGradient} estimator. It iteratively fits models, while correcting violations of the specified constraint from previous iterations. In the end, a randomized ensemble is formed, to then conduct predictions. We employ two constraints: Demographic Parity (DP) and Equalized Odds (EO). Yet, since the online model is incompatible with the \texttt{ExponentiatedGradient} estimator, we adapt a \texttt{xgboost}~\cite{chen2016xgboost} classifier for a streaming context. It implements incremental updates, by storing a booster cache after the first fit, to which we iteratively append new trees at each \texttt{update\_interval}, without having to fully re-train. We then wrap the ensemble on a \texttt{ExponentiatedGradient} model.

    \section{Results}
    \label{sec:results}

    We compute composite scores regarding performance (i.e., combining ROC AUC, accuracy, precision, recall, and log loss), and fairness (i.e., combining SPD and EOD), to conduct rank analyses in each setting. In Table~\ref{tab:summary_ranks_combined}, the counts of all best and second-best ranks attained by each method in both pipelines, are shown. Estimations of performance disaggregated by groups, are also shown in Table~\ref{tab:pergroup-results-static} and Table~\ref{tab:pergroup-results-online}, for the offline and online pipelines respectively. Moreover, overall results, alongside the code used in the experiments, are all available online.~\footnote{\href{https://github.com/ricardoinaciopt/bias_mitigation/tree/main/results_tables}{https://github.com/ricardoinaciopt/bias\_mitigation/tree/main/results\_tables}}
    
    \begin{table}[h]
      \centering
      \footnotesize
      \caption{Counts of all 1st and 2nd ranks each mitigation technique achieved, in performance and fairness, across all scenarios. Bold denotes the most recurrent method.}
      \label{tab:summary_ranks_combined}
      \begin{tabular}{llcc@{\hskip 10pt}cc}
        \toprule
        \multicolumn{1}{c}{\textbf{Pipeline}} & \multicolumn{1}{c}{\textbf{Mitigation Technique}} 
          & \multicolumn{2}{c}{\textbf{Performance}} 
          & \multicolumn{2}{c}{\textbf{Fairness}} \\
        \cmidrule(lr){3-4} \cmidrule(lr){5-6}
          &  & 1st & 2nd & 1st & 2nd \\
        \midrule
        \multirow{5}{*}{Offline}
          & Baseline (None)                            &  1 &  \best{9}   &  0 &  1 \\
          & Auto Reweighing                 &  0 &  0          &  0 &  \best{9} \\
          & Manual Reweighing               &  1 &  5          &  0 &  0 \\
          & Demographic Parity Constraint   &  0 &  1          &  \best{16} &  0 \\
          & Equalized Odds Constraint       &  \best{14} &  1  &  0 &  6 \\
        \midrule
        \multirow{5}{*}{Online}
          & Baseline (None)                            &  \best{6} & 5   &  0 &  0 \\
          & Auto Reweighing                 &  \best{6} &  4           &  0 &  0 \\
          & Manual Reweighing               &  4        & \best{7}     &  0 &  0 \\
          & Demographic Parity Constraint   &  0        &  0           &  \best{15} & 1 \\
          & Equalized Odds Constraint       &  0        &  0           & 1 & \best{15} \\
        \bottomrule
      \end{tabular}
    \end{table}
  
    \subsection{Baselines}
    
    These scenarios represent the basis for comparison to all mitigation techniques, as to gauge their respective effect across performance and fairness. 

    \paragraph{\textbf{Offline pipeline.}} In 9 of the 16 scenarios, the second-best rank in performance was attained, indicating that even as is, the classifier is capable to approve truly qualified loans. However, this aggregated performance conceals disparities, highlighted in group-wise metrics, which are considerably lower for group B. Only in the scenario where no explicit biases were introduced (i.e., \texttt{lbl} = 0.0, \texttt{rep} = 0.5), results were fairly balanced across groups, except for \texttt{recall}, where the gap highlights that class-inherent biases might lead to the negligence of group B entities. Considering fairness, as it was expected, some of the worst overall results were attained, given that all biases exert effect without constraints.

    \paragraph{\textbf{Online pipeline.}} The baselines were able to achieve best and second-best performance ranks 6 and 5 times out of 16, respectively. Group-based discrepancies also surface across most performance metrics, more prominently when the prejudice (\texttt{lbl}) bias is large. Fairness is likewise poor for this setting, since in no scenario it attained best or second-best ranks. Both metrics boast very large values, all pointing in the direction of the privileged group.
        
    \subsection{Automatic Reweighing}
    
    A pre-processing technique that re-weights each instance considering both its group, and the respective target label, before carrying out predictions. Since it does not change the underlying algorithm, it is agnostic and configuration-free.

    \paragraph{\textbf{Offline pipeline.}} The performance of the classifier degraded in all scenarios, relative to the baselines, never attaining best or second-best ranks. Across most group-wise performance metrics, values corresponding to group B are considerably lower in all scenarios, having the gap increase alongside the level of imbalance and prejudice (\texttt{rep} and \texttt{lbl}) biases. Regarding fairness, it was promising, as it achieved the second-best rank in 9 of 16 cases, making it a viable option considering its simplicity and broad compatibility. It should be noted that, although fairness improved relative to the baseline in most cases, the improvement was not substantial. Both SPD and EOD were still slightly far from zero.  

    \paragraph{\textbf{Online pipeline.}} In terms of relative performance, it was analogous to the baseline as it also ranked first 6 out of 16 times. However, absolute values highlight that it only outperformed the baseline variant in half of the scenarios, and when it did, the improvement was not substantial. Nonetheless, although most of the group-wise disparities were still present and pronounced, the gaps in performance were occasionally abridged. Considering overall fairness, although it never attained best or second-best ranks, it was able to surpass the baseline in most cases, making it a viable approach to improve unfairness, in a lightweight manner that does not decrease performance substantially. 

    \subsection{Manual Reweighing}

    Here, we directly favour the protected group, while disfavouring the privileged one $(A:0.5, B:1.5)$, to assess if this simple boost is able to counteract biases.
    
    \paragraph{\textbf{Offline pipeline.}} Even though performance was analogous to the baseline in most cases, it was still able to improve it occasionally, mostly for group B, as seen in the group-wise metrics, showcasing how easily the classifier can be skewed. Although it attained best and second-best performance ranks 1 and 5 out of 16 times respectively, it could not consistently surpass the baseline. On overall fairness, it never attained best or second-best ranks. Interestingly, group-based structural biases still exert some effect, even when group B is pronounced.

    \paragraph{\textbf{Online pipeline.}} It only surpassed the baseline in less than half of the scenarios, where the improvement was mostly marginal. Group-wise performance, although not highly discrepant, still favours group A instances, evident by the consistently higher approval rate. Overall fairness remarkably improved, highlighted by EOD, which decreased in all but one case, being almost always fairer than the baseline. Furthermore, in cases without any prejudice bias ($\texttt{lbl} = 0$), the outcomes closely approached 0, showcasing viability in modestly biased scenarios.

    \subsection{Constrained ExponenetiatedGradient Estimator}
    
    Until this point, mitigation was applied through pre-processing, which means the classifiers were not directly affected. Yet, although results have been promising in both fairness and performance, improvements have not been significant. To that end, an in-processing technique is now employed (i.e., at the algorithm level), namely the \texttt{ExponetiatedGradient} estimator, by applying both \texttt{Equalized Odds} and \texttt{Demographic Parity} constraints.

    \subsubsection{Demographic Parity Mitigation.}

    This constraint enforces the classifier to carry predictions that are statistically independent of the sensitive feature, which entails that positive outcomes cannot be influenced by the group of each entity.
    
    \paragraph{\textbf{Offline pipeline.}}  Effectiveness is evident on fairness, achieving the best rank in all 16 scenarios. In most cases, the approval rate across groups was approximated successfully, only slightly diverging on the cases with the strongest prejudice bias (\texttt{lbl}). The effect on SPD specifically was expected, driving it in most cases very close to 0, aptly meeting the criteria of the constraint. Furthermore, EOD also severely improved, even without constraining to model to it.

    \paragraph{\textbf{Online pipeline.}} It never attained best or second best ranks in overall performance, not even surpassing the baseline once. Nonetheless, in cases where both \texttt{lbl} and \texttt{rep} were high, \texttt{recall} and \texttt{approval rates} significantly improved for group B , indicating that while it generally hampered the classifier, it led to slightly fairer outcomes. In fairness, it was also the best, as it ranked first in 15 out of 16 cases, being very effective in approaching both SPD and EOD to 0.

    \subsubsection{Equalized Odds Mitigation.}
    
    Besides enforcing the independence of results from group belonging, this constraint also takes into account the true labels, which entails considering the real outcomes as well.

    \paragraph{\textbf{Offline pipeline.}} It not only attained great performance overall, being ranked best in 14 of 16 scenarios, but also satisfactory fairness, being 6 out of 16 times the second-best. However, regarding both fairness metrics, although it was often able to surpass the baselines, improvement was mostly negligible when it did. Furthermore, given the high discrepancies in group-wise metrics, both groups could lie on misaligned ROC curves that might not intersect, based on large recall, precision, and approval-rate gaps. Hence, in case no (TPR, FPR) threshold exists across both group-specific curves, their lower envelope must be leveraged instead, effectively sacrificing the performance of group A, in order to meet the one from group B, as it is observed in the outcomes.

    \paragraph{\textbf{Online pipeline.}} It never attained best or second-best overall performance ranks, and never surpassed baseline values. Yet, group-wise approval rates showcase a considerable improvement on those specific to group B. Considering fairness, in 15 out of 16 scenarios it led to a substantial improvement on both metrics, attaining second-best rank on those instances. This makes it a viable runner up, when predictive performance is not crucial.

    \section{Mitigation Impact on Predictions via Shapley Values}
    \label{sec:xai}

    In order to assess the effects of mitigation on classifiers, we analyse how it shifts the way features are leveraged for predictions. Then, we explain a random instance in each modality, considering both the baseline and the most effective mitigation (i.e., demographic parity constraint), in the least and most biased settings. Although the randomness inherent to the mitigation may introduce some variability in the results, the observed patterns hold for most cases.
    
    Overall, individual contributions (i.e., node size) were spread across a broader feature set, which in turn reduced the prevalence of hubs (i.e., disproportionality large, highly predictive nodes). Also, when a small set of interactions (i.e., thick edges) boasted most of the influence, the mitigation also spread it across a broader set of pairs. However, when these were highly predictive, the method could not reduce their impact without also hindering performance substantially, bellow an acceptable threshold (i.e., defined by the slack parameter~\cite{weerts2023fairlearn}). Hence, some networks still boast strong dominant interactions post-mitigation.  Moreover, redundant interactions (i.e., blue edges), were typically left unchanged, more prominently in highly biased scenarios, being sometimes even emphasized, given that these do not strongly impact fairness. For analogous reasons, the contribution of a node or edge was occasionally flipped to negative.

    \subsection{Offline Classifier}

\begin{figure}[h]
        \centering
        \includegraphics[width=\linewidth]{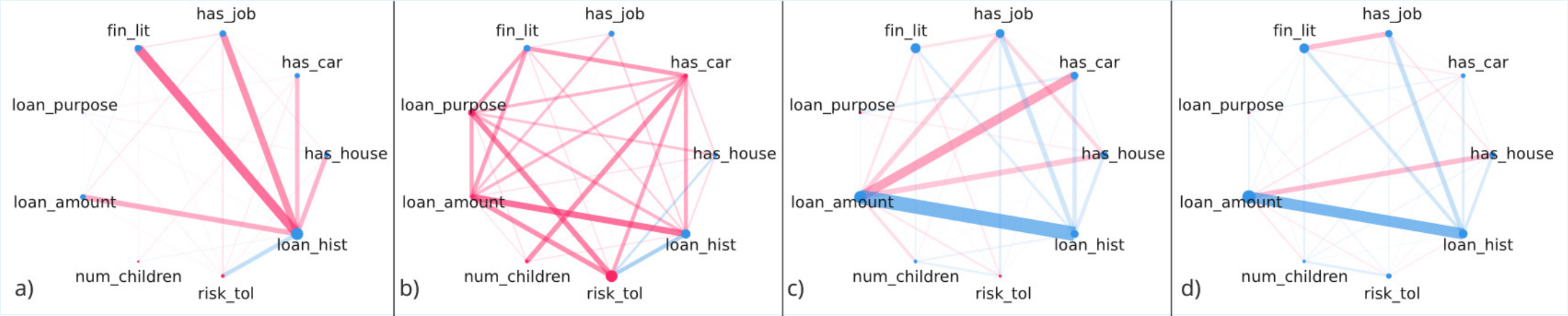}
        \caption{Shapley network plots for the least (\texttt{a)}  and \texttt{b)}) and most (\texttt{c)} and \texttt{d)}) explicitly biased scenarios in the offline modality. These show the baselines alongside the models constrained by demographic parity.}
        \label{fig:offline-shap}
    \end{figure}

    In the least biased scenario (\texttt{a)} to \texttt{b)}), the \texttt{loan\_hist} feature node is clearly a hub, since it is the largest, and most strong edges include it. Thus, the mitigation did not only spread its weight throughout the network, as it also increased predictive impact of other features and interactions. In the highly biased scenario (\texttt{c)} to \texttt{d)}), although the weight of most hubs and related edges was largely distilled (e.g., \texttt{loan\_amount}), the method was not able to diffuse the strongest interaction (c.f. \texttt{d)} in Figure~\ref{fig:offline-shap}), implying that it is an overly influential predictor.
  
    \subsection{Online Classifier}
    
\begin{figure}[h]
        \centering
        \includegraphics[width=\linewidth]{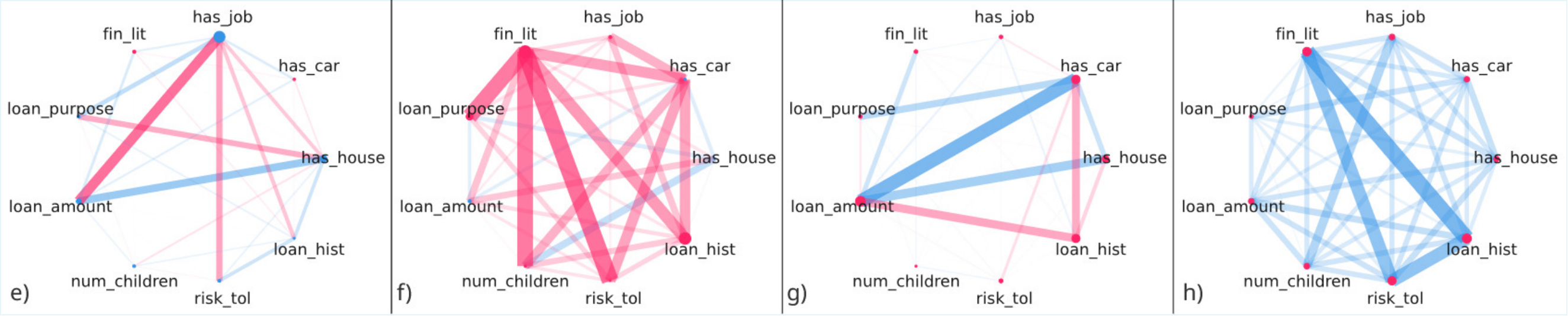}
        \caption{Shapley network plots for the least (\texttt{e)}  and \texttt{f)}) and most (\texttt{g)} and \texttt{h)}) explicitly biased scenarios in the online modality. These show the baselines alongside the models constrained by demographic parity.}
        \label{fig:online-shap}
    \end{figure}

     Across bias levels, results were more stable overall, with most of the previously documented patterns frequently resurfacing. This might be attributed to the streaming nature of the model, which applies constraints to the mini-batches as it receives them, basically enforcing subset-level fairness. In the less biased setting (\texttt{e)} to \texttt{f)} in Figure~\ref{fig:online-shap}), the same diffusion of relevance from hubs to the remaining nodes of the network was observed, which led to more balanced contributions. In more biased settings, such as \texttt{g)} to \texttt{h)}, a preference for condensed, but redundant (i.e., blue) networks was perceived, given that these are not as impactful for unfairness. However, in the cases where diluting highly predictive features or strong interactions was challenging, it was nonetheless able spread contributions across a broader feature subset. Particularly in this modality, mitigations regularly led to nearly complete interaction graphs.
    
    \section{Conclusion}
	\label{sec:conclusion}

    In this work, we showcased how bias concealed in the processes of a simulated society, can be embedded in data gathered from them. Later, such patterns were shown to be reinforced by classifiers trained on it. An ABM that depicts both a loan approval process, and transactions across peers with diverse levels of bias, was developed to generate said data. Subsequently, different mitigation techniques were applied, demonstrating how methods of varying complexity could alleviate the negative effects of disparities. Our findings highlight that mitigation approaches, although generally effective, always entail some sort of loss in predictive performance. 
    
    Simple pre-processing reweighing approaches, either automatically or manually, although not very effective, are often able to reach a balanced trade-off, while being both lightweight and method-agnostic. In-processing methods, such as the constrained \texttt{ExponentiatedGradient} estimator, were often aggressive approaches that either improve performance or fairness exceptionally, but not in tandem. Particularly, \texttt{demographic parity} led to the fairest outcomes, whereas \texttt{equalized odds} consistently attained satisfactory fairness results, while not degrading performance as much.

    Although an example is shown through a loan application process, the actual contribution is the underlying framework for the generation of synthetic datasets through an ABM with controllable bias injection, which can be applied to any setting. Shapley values are then used to assess the effects of mitigations in classifiers. We found that generally, interaction networks get diluted across most features, leading to more balanced contributions. In the presence of highly predictive nodes or edges, which cannot be attenuated without severely hurting performance, it could nonetheless spread them across broader feature subsets.
    
    \bibliographystyle{splncs04}
    \bibliography{mybibliography}

\begin{thebibliography}{10}
\providecommand{\url}[1]{\texttt{#1}}
\providecommand{\urlprefix}{URL }
\providecommand{\doi}[1]{https://doi.org/#1}

\bibitem{agarwal2018reductions}
Agarwal, A., Beygelzimer, A., Dud{\'\i}k, M., Langford, J., Wallach, H.: A reductions approach to fair classification. In: International conference on machine learning. pp. 60--69. PMLR (2018)

\bibitem{aif360-oct-2018}
Bellamy, R.K.E., Dey, K., Hind, M., Hoffman, S.C., Houde, S., Kannan, K., Lohia, P., Martino, J., Mehta, S., Mojsilovic, A., Nagar, S., Ramamurthy, K.N., Richards, J., Saha, D., Sattigeri, P., Singh, M., Varshney, K.R., Zhang, Y.: {AI Fairness} 360: An extensible toolkit for detecting, understanding, and mitigating unwanted algorithmic bias (Oct 2018), \url{https://arxiv.org/abs/1810.01943}

\bibitem{doi:10.1073/pnas.082080899}
Bonabeau, E.: Agent-based modeling: Methods and techniques for simulating human systems. Proceedings of the National Academy of Sciences  \textbf{99}(suppl\_3),  7280--7287 (2002). \doi{10.1073/pnas.082080899}, \url{https://www.pnas.org/doi/abs/10.1073/pnas.082080899}

\bibitem{chen2016xgboost}
Chen, T., Guestrin, C.: Xgboost: A scalable tree boosting system. In: Proceedings of the 22nd acm sigkdd international conference on knowledge discovery and data mining. pp. 785--794 (2016)

\bibitem{de2014agent}
De~Marchi, S., Page, S.E.: Agent-based models. Annual Review of political science  \textbf{17}(1),  1--20 (2014)

\bibitem{epstein1996growing}
Epstein, J.M., Axtell, R.: Growing artificial societies: social science from the bottom up. Brookings Institution Press (1996)

\bibitem{goethals2024beyond}
Goethals, S., Calders, T., Martens, D.: Beyond accuracy-fairness: Stop evaluating bias mitigation methods solely on between-group metrics  (2024)

\bibitem{hagberg2008exploring}
Hagberg, A., Swart, P.J., Schult, D.A.: Exploring network structure, dynamics, and function using networkx. Tech. rep., Los Alamos National Laboratory (LANL), Los Alamos, NM (United States) (2008)

\bibitem{haixiang2017learning}
Haixiang, G., Yijing, L., Shang, J., Mingyun, G., Yuanyue, H., Bing, G.: Learning from class-imbalanced data: Review of methods and applications. Expert systems with applications  \textbf{73},  220--239 (2017)

\bibitem{hardt2016equality}
Hardt, M., Price, E., Srebro, N.: Equality of opportunity in supervised learning. Advances in neural information processing systems  \textbf{29} (2016)

\bibitem{hoens2013imbalanced}
Hoens, T.R., Chawla, N.V.: Imbalanced datasets: from sampling to classifiers. Imbalanced learning: Foundations, algorithms, and applications pp. 43--59 (2013)

\bibitem{hoi2021online}
Hoi, S.C., Sahoo, D., Lu, J., Zhao, P.: Online learning: A comprehensive survey. Neurocomputing  \textbf{459},  249--289 (2021)

\bibitem{huang2022evaluation}
Huang, J., Galal, G., Etemadi, M., Vaidyanathan, M.: Evaluation and mitigation of racial bias in clinical machine learning models: scoping review. JMIR Medical Informatics  \textbf{10}(5),  e36388 (2022)

\bibitem{kamiran2012data}
Kamiran, F., Calders, T.: Data preprocessing techniques for classification without discrimination. Knowledge and information systems  \textbf{33}(1),  1--33 (2012)

\bibitem{klinkenberg2004learning}
Klinkenberg, R.: Learning drifting concepts: Example selection vs. example weighting. Intelligent data analysis  \textbf{8}(3),  281--300 (2004)

\bibitem{lundberg2017unified}
Lundberg, S.M., Lee, S.I.: A unified approach to interpreting model predictions. Advances in neural information processing systems  \textbf{30} (2017)

\bibitem{mehrabi2021survey}
Mehrabi, N., Morstatter, F., Saxena, N., Lerman, K., Galstyan, A.: A survey on bias and fairness in machine learning. ACM computing surveys (CSUR)  \textbf{54}(6),  1--35 (2021)

\bibitem{montiel2021river}
Montiel, J., Halford, M., Mastelini, S.M., Bolmier, G., Sourty, R., Vaysse, R., Zouitine, A., Gomes, H.M., Read, J., Abdessalem, T., et~al.: River: machine learning for streaming data in python  (2021)

\bibitem{muschalik2024shapiq}
Muschalik, M., Baniecki, H., Fumagalli, F., Kolpaczki, P., Hammer, B., H\"{u}llermeier, E.: shapiq: Shapley interactions for machine learning. In: The Thirty-eight Conference on Neural Information Processing Systems Datasets and Benchmarks Track (2024), \url{https://openreview.net/forum?id=knxGmi6SJi}

\bibitem{ntoutsi2020bias}
Ntoutsi, E., Fafalios, P., Gadiraju, U., Iosifidis, V., Nejdl, W., Vidal, M.E., Ruggieri, S., Turini, F., Papadopoulos, S., Krasanakis, E., et~al.: Bias in data-driven artificial intelligence systems—an introductory survey. Wiley Interdisciplinary Reviews: Data Mining and Knowledge Discovery  \textbf{10}(3),  e1356 (2020)

\bibitem{perez2018review}
P{\'e}rez-S{\'a}nchez, B., Fontenla-Romero, O., Guijarro-Berdi{\~n}as, B.: A review of adaptive online learning for artificial neural networks. Artificial Intelligence Review  \textbf{49},  281--299 (2018)

\bibitem{savani2020intra}
Savani, Y., White, C., Govindarajulu, N.S.: Intra-processing methods for debiasing neural networks. Advances in neural information processing systems  \textbf{33},  2798--2810 (2020)

\bibitem{schelling1971}
Schelling, T.C.: Dynamic models of segregation. Journal of Mathematical Sociology  \textbf{1}(2),  143 -- 186 (1971)

\bibitem{wang2021directional}
Wang, A., Russakovsky, O.: Directional bias amplification. In: International Conference on Machine Learning. pp. 10882--10893. PMLR (2021)

\bibitem{wang2023overwriting}
Wang, A., Russakovsky, O.: Overwriting pretrained bias with finetuning data. In: Proceedings of the IEEE/CVF International Conference on Computer Vision. pp. 3957--3968 (2023)

\bibitem{weerts2023fairlearn}
Weerts, H., Dudík, M., Edgar, R., Jalali, A., Lutz, R., Madaio, M.: Fairlearn: Assessing and improving fairness of ai systems (2023), \url{http://jmlr.org/papers/v24/23-0389.html}

\bibitem{zhang2020deterministic}
Zhang, H., Zhang, Y., Zhu, S., Xu, D.: Deterministic convergence of complex mini-batch gradient learning algorithm for fully complex-valued neural networks. Neurocomputing  \textbf{407},  185--193 (2020)

\end{thebibliography}

   \appendix
   \begin{table*}[t]
    \centering
    \footnotesize
    \caption{Per–group results (accuracy, recall, precision and approval‐rate disaggregated by groups A and B) for the offline pipeline.}
    \label{tab:pergroup-results-static}
    \resizebox{0.8\textwidth}{!}{%
    \begin{tabular}{lcccccccc}
    \toprule
    Variant &
    A Acc.\,$(\uparrow)$ & B Acc.\,$(\uparrow)$ &
    A Rec.\,$(\uparrow)$ & B Rec.\,$(\uparrow)$ &
    A Prec.\,$(\uparrow)$ & B Prec.\,$(\uparrow)$ &
    A AppRate\,$(\uparrow)$ & B AppRate\,$(\uparrow)$\\
    \midrule
    \multicolumn{9}{c}{\textbf{rep = 0.5 | lbl = 0.0 | A support = 632 | B support = 579}}\\
    none                         & \best{0.935} & \best{0.967} & \best{0.954} & 0.830 & 0.942 & \best{0.976} & \best{0.630} & 0.147\\
    reweight\_auto                & \second{0.932} & 0.957 & 0.939 & \second{0.870} & \second{0.951} & 0.879 & 0.614 & \second{0.171}\\
    reweight\_manual\_A0.5\_B1.5  & 0.921 & 0.960 & \second{0.941} & 0.800 & 0.932 & \second{0.964} & \second{0.628} & 0.143\\
    mitigator\_demographic\_parity & 0.894 & 0.611 & 0.847 & \best{0.910} & \best{0.979} & 0.296 & 0.538 & \best{0.530}\\
    mitigator\_equalized\_odds     & 0.929 & \second{0.962} & 0.939 & 0.850 & 0.946 & 0.924 & 0.617 & 0.159\\
    \midrule
    \multicolumn{9}{c}{\textbf{rep = 0.5 | lbl = 0.4 | A support = 616 | B support = 605}}\\
    none                         & \second{0.920} & \best{0.954} & \best{0.948} & 0.568 & 0.941 & \best{0.636} & \second{0.718} & 0.055\\
    reweight\_auto                & \best{0.922} & 0.929 & \best{0.948} & \second{0.676} & 0.943 & 0.446 & 0.716 & \second{0.093}\\
    reweight\_manual\_A0.5\_B1.5  & 0.906 & \best{0.954} & 0.938 & 0.595 & 0.930 & \second{0.629} & \best{0.719} & 0.058\\
    mitigator\_demographic\_parity & 0.846 & 0.383 & 0.800 & \best{0.838} & \best{0.980} & 0.078 & 0.581 & \best{0.658}\\
    mitigator\_equalized\_odds     & 0.909 & 0.944 & 0.918 & 0.568 & \second{0.953} & 0.538 & 0.687 & 0.064\\
    \midrule
    \multicolumn{9}{c}{\textbf{rep = 0.5 | lbl = 0.5 | A support = 644 | B support = 576}}\\
    none                         & \best{0.932} & 0.955 & \best{0.965} & \second{0.677} & 0.940 & 0.568 & \best{0.730} & \second{0.064}\\
    reweight\_auto                & 0.904 & \second{0.960} & 0.891 & 0.581 & \best{0.971} & \second{0.643} & 0.652 & 0.049\\
    reweight\_manual\_A0.5\_B1.5  & 0.915 & \best{0.972} & 0.921 & 0.645 & 0.957 & \best{0.800} & 0.685 & 0.043\\
    mitigator\_demographic\_parity & 0.846 & 0.451 & 0.808 & \best{0.903} & \second{0.971} & 0.082 & 0.592 & \best{0.592}\\
    mitigator\_equalized\_odds     & \second{0.925} & \second{0.960} & \second{0.932} & \second{0.677} & 0.962 & 0.618 & \second{0.689} & 0.059\\
    \midrule
    \multicolumn{9}{c}{\textbf{rep = 0.5 | lbl = 0.6 | A support = 618 | B support = 597}}\\
    none                         & 0.896 & \second{0.940} & 0.928 & 0.267 & 0.928 & \second{0.364} & \second{0.718} & 0.037\\
    reweight\_auto                & \best{0.906} & 0.925 & \best{0.948} & \second{0.567} & 0.923 & 0.347 & \best{0.738} & \second{0.082}\\
    reweight\_manual\_A0.5\_B1.5  & 0.890 & \best{0.951} & 0.917 & 0.300 & 0.929 & \best{0.529} & 0.709 & 0.028\\
    mitigator\_demographic\_parity & 0.366 & 0.787 & 0.119 & \best{0.767} & \best{0.981} & 0.161 & 0.087 & \best{0.240}\\
    mitigator\_equalized\_odds     & \best{0.906} & 0.938 & \second{0.935} & 0.300 & \second{0.935} & 0.360 & \second{0.718} & 0.042\\
    \midrule
    \multicolumn{9}{c}{\textbf{rep = 0.6 | lbl = 0.0 | A support = 771 | B support = 455}}\\
    none                         & 0.917 & \best{0.949} & 0.925 & 0.816 & \second{0.933} & \best{0.873} & 0.585 & 0.156\\
    reweight\_auto                & 0.913 & 0.941 & 0.919 & \second{0.842} & 0.933 & 0.810 & 0.581 & \second{0.174}\\
    reweight\_manual\_A0.5\_B1.5  & \best{0.926} & \second{0.947} & \best{0.952} & 0.829 & 0.925 & \second{0.851} & \best{0.607} & 0.163\\
    mitigator\_demographic\_parity & 0.913 & 0.596 & 0.914 & \best{0.921} & \best{0.937} & 0.282 & 0.576 & \best{0.545}\\
    mitigator\_equalized\_odds     & \second{0.925} & \second{0.947} & \second{0.947} & \second{0.842} & 0.927 & 0.842 & \second{0.603} & 0.167\\
    \midrule
    \multicolumn{9}{c}{\textbf{rep = 0.6 | lbl = 0.4 | A support = 771 | B support = 463}}\\
    none                         & 0.904 & \second{0.937} & 0.905 & 0.351 & \best{0.957} & \second{0.722} & 0.668 & 0.039\\
    reweight\_auto                & \best{0.925} & 0.922 & 0.945 & 0.541 & 0.948 & 0.513 & 0.704 & \second{0.084}\\
    reweight\_manual\_A0.5\_B1.5  & 0.909 & \best{0.942} & 0.919 & 0.432 & \second{0.951} & \best{0.727} & 0.684 & 0.048\\
    mitigator\_demographic\_parity & 0.917 & 0.335 & \second{0.949} & \best{0.946} & 0.935 & 0.103 & \second{0.717} & \best{0.737}\\
    mitigator\_equalized\_odds     & \best{0.925} & 0.933 & \best{0.960} & \second{0.595} & 0.936 & 0.579 & \best{0.725} & 0.082\\
    \midrule
    \multicolumn{9}{c}{\textbf{rep = 0.6 | lbl = 0.5 | A support = 776 | B support = 461}}\\
    none                         & \second{0.930} & \second{0.954} & 0.913 & \second{0.484} & \second{0.984} & \second{0.750} & 0.635 & 0.043\\
    reweight\_auto                & 0.929 & 0.948 & 0.917 & \second{0.484} & 0.978 & 0.652 & 0.642 & 0.050\\
    reweight\_manual\_A0.5\_B1.5  & 0.912 & \best{0.957} & 0.885 & 0.419 & \best{0.985} & \best{0.867} & 0.615 & 0.033\\
    mitigator\_demographic\_parity & \second{0.930} & 0.282 & \best{0.951} & \best{0.871} & 0.947 & 0.076 & \best{0.687} & \best{0.768}\\
    mitigator\_equalized\_odds     & \best{0.936} & 0.946 & \second{0.940} & \second{0.484} & 0.965 & 0.625 & \second{0.666} & \second{0.052}\\
    \midrule
    \multicolumn{9}{c}{\textbf{rep = 0.6 | lbl = 0.6 | A support = 763 | B support = 461}}\\
    none                         & \second{0.933} & \best{0.946} & 0.970 & \second{0.450} & \second{0.941} & \best{0.391} & 0.754 & 0.050\\
    reweight\_auto                & 0.928 & 0.920 & 0.970 & \second{0.450} & 0.934 & 0.257 & 0.759 & \second{0.076}\\
    reweight\_manual\_A0.5\_B1.5  & 0.924 & \best{0.946} & \best{0.980} & \second{0.450} & 0.921 & \best{0.391} & \best{0.779} & 0.050\\
    mitigator\_demographic\_parity & 0.932 & 0.258 & \second{0.973} & \best{0.900} & 0.936 & 0.050 & \second{0.760} & \best{0.777}\\
    mitigator\_equalized\_odds     & \best{0.934} & \best{0.946} & 0.971 & \second{0.450} & \best{0.941} & \best{0.391} & 0.755 & 0.050\\
    \midrule
    \multicolumn{9}{c}{\textbf{rep = 0.7 | lbl = 0.0 | A support = 881 | B support = 338}}\\
    none                         & \best{0.933} & 0.938 & \best{0.950} & 0.806 & 0.938 & 0.847 & \second{0.604} & 0.175\\
    reweight\_auto                & 0.925 & \best{0.947} & 0.930 & 0.806 & \best{0.944} & \best{0.893} & 0.587 & 0.166\\
    reweight\_manual\_A0.5\_B1.5  & 0.927 & 0.944 & 0.937 & 0.806 & \second{0.941} & \second{0.877} & 0.594 & 0.169\\
    mitigator\_demographic\_parity & \second{0.930} & 0.571 & 0.943 & \best{0.935} & 0.939 & 0.291 & 0.598 & \best{0.589}\\
    mitigator\_equalized\_odds     & \second{0.930} & \best{0.947} & \second{0.949} & \second{0.839} & 0.934 & 0.867 & \best{0.605} & \second{0.178}\\
    \midrule
    \multicolumn{9}{c}{\textbf{rep = 0.7 | lbl = 0.4 | A support = 890 | B support = 349}}\\
    none                         & \best{0.939} & \second{0.943} & \best{0.965} & 0.633 & 0.949 & \second{0.679} & \best{0.709} & 0.080\\
    reweight\_auto                & \best{0.939} & 0.926 & 0.955 & \second{0.667} & \best{0.958} & 0.556 & 0.696 & \second{0.103}\\
    reweight\_manual\_A0.5\_B1.5  & 0.936 & 0.940 & \second{0.960} & 0.633 & 0.949 & 0.655 & \second{0.706} & 0.083\\
    mitigator\_demographic\_parity & 0.936 & 0.384 & 0.958 & \best{0.867} & 0.950 & 0.110 & 0.703 & \best{0.679}\\
    mitigator\_equalized\_odds     & 0.938 & \best{0.946} & 0.953 & 0.633 & \second{0.958} & \best{0.704} & 0.694 & 0.077\\
    \midrule
    \multicolumn{9}{c}{\textbf{rep = 0.7 | lbl = 0.5 | A support = 887 | B support = 360}}\\
    none                         & \best{0.939} & 0.908 & \best{0.977} & \second{0.588} & 0.942 & 0.278 & \best{0.754} & \second{0.100}\\
    reweight\_auto                & 0.933 & \second{0.928} & 0.946 & 0.471 & \best{0.962} & 0.320 & 0.715 & 0.069\\
    reweight\_manual\_A0.5\_B1.5  & 0.928 & \best{0.947} & 0.938 & 0.529 & \second{0.962} & \best{0.450} & 0.709 & 0.056\\
    mitigator\_demographic\_parity & 0.937 & 0.192 & \second{0.974} & \best{0.941} & 0.942 & 0.052 & \second{0.752} & \best{0.850}\\
    mitigator\_equalized\_odds     & \best{0.939} & \second{0.928} & 0.961 & \second{0.588} & 0.955 & \second{0.345} & 0.732 & 0.081\\
    \midrule
    \multicolumn{9}{c}{\textbf{rep = 0.7 | lbl = 0.6 | A support = 908 | B support = 344}}\\
    none                         & 0.909 & \second{0.936} & \second{0.953} & 0.533 & 0.923 & \second{0.348} & \second{0.744} & 0.067\\
    reweight\_auto                & \second{0.912} & 0.904 & \best{0.963} & \second{0.600} & 0.918 & 0.250 & \best{0.757} & \second{0.105}\\
    reweight\_manual\_A0.5\_B1.5  & 0.907 & \best{0.956} & 0.939 & 0.467 & 0.933 & \best{0.500} & 0.726 & 0.041\\
    mitigator\_demographic\_parity & \best{0.924} & 0.363 & \second{0.953} & \best{0.667} & \best{0.943} & 0.045 & 0.729 & \best{0.651}\\
    mitigator\_equalized\_odds     & 0.910 & \second{0.936} & 0.940 & 0.533 & \second{0.935} & \second{0.348} & 0.726 & 0.067\\
    \midrule
    \multicolumn{9}{c}{\textbf{rep = 0.8 | lbl = 0.0 | A support = 1030 | B support = 216}}\\
    none                         & 0.911 & \best{0.944} & \second{0.926} & \second{0.818} & 0.923 & \best{0.900} & \second{0.589} & 0.185\\
    reweight\_auto                & \second{0.920} & 0.940 & 0.924 & \second{0.818} & \best{0.939} & 0.878 & 0.578 & \second{0.190}\\
    reweight\_manual\_A0.5\_B1.5  & 0.914 & 0.940 & 0.922 & 0.795 & 0.930 & 0.897 & 0.583 & 0.181\\
    mitigator\_demographic\_parity & \best{0.925} & 0.736 & \best{0.940} & \best{0.909} & \second{0.933} & 0.430 & \best{0.592} & \best{0.431}\\
    mitigator\_equalized\_odds     & 0.916 & \best{0.944} & 0.924 & \second{0.818} & 0.932 & \best{0.900} & 0.583 & 0.185\\
    \midrule
    \multicolumn{9}{c}{\textbf{rep = 0.8 | lbl = 0.4 | A support = 1017 | B support = 234}}\\
    none                         & 0.942 & \second{0.936} & 0.946 & 0.533 & \best{0.971} & \second{0.500} & 0.687 & 0.068\\
    reweight\_auto                & 0.940 & 0.919 & 0.946 & 0.600 & 0.969 & 0.409 & 0.689 & 0.094\\
    reweight\_manual\_A0.5\_B1.5  & \second{0.947} & \best{0.944} & 0.955 & 0.467 & \second{0.969} & \best{0.583} & 0.696 & 0.051\\
    mitigator\_demographic\_parity & 0.946 & 0.333 & \best{0.965} & \best{0.800} & 0.959 & 0.073 & \best{0.711} & \best{0.705}\\
    mitigator\_equalized\_odds     & \best{0.952} & 0.932 & \best{0.965} & \second{0.733} & 0.967 & 0.478 & \second{0.705} & \second{0.098}\\
    \midrule
    \multicolumn{9}{c}{\textbf{rep = 0.8 | lbl = 0.5 | A support = 1007 | B support = 234}}\\
    none                         & \best{0.948} & 0.923 & \best{0.972} & 0.222 & 0.956 & 0.154 & \best{0.725} & 0.056\\
    reweight\_auto                & 0.929 & \second{0.932} & 0.940 & 0.222 & \best{0.960} & 0.182 & 0.698 & 0.047\\
    reweight\_manual\_A0.5\_B1.5  & 0.935 & \best{0.953} & 0.951 & 0.222 & \second{0.958} & \best{0.333} & 0.708 & 0.026\\
    mitigator\_demographic\_parity & 0.939 & 0.350 & 0.965 & \best{0.889} & 0.951 & 0.050 & \second{0.724} & \best{0.679}\\
    mitigator\_equalized\_odds     & \second{0.946} & 0.923 & \second{0.968} & \second{0.444} & 0.957 & \second{0.235} & 0.721 & \second{0.073}\\
    \midrule
    \multicolumn{9}{c}{\textbf{rep = 0.8 | lbl = 0.6 | A support = 1016 | B support = 222}}\\
    none                         & \best{0.932} & \second{0.896} & \second{0.972} & 0.286 & 0.938 & \second{0.235} & \second{0.756} & 0.077\\
    reweight\_auto                & 0.929 & 0.815 & \best{0.989} & \second{0.571} & 0.920 & 0.186 & \best{0.784} & \second{0.194}\\
    reweight\_manual\_A0.5\_B1.5  & \second{0.930} & \best{0.910} & 0.962 & 0.357 & \best{0.943} & \best{0.312} & 0.744 & 0.072\\
    mitigator\_demographic\_parity & 0.927 & 0.374 & 0.966 & \best{0.857} & 0.936 & 0.081 & 0.753 & \best{0.671}\\
    mitigator\_equalized\_odds     & 0.925 & 0.874 & 0.961 & 0.429 & \second{0.938} & 0.231 & 0.747 & 0.117\\
    \bottomrule
    \end{tabular}}%
    \end{table*}

    \begin{table*}[t]
    \centering
    \footnotesize
    \caption{Per–group results (accuracy, recall, precision and approval-rate disaggregated by groups A and B) for the online pipeline.}
    \label{tab:pergroup-results-online}
    \resizebox{0.8\textwidth}{!}{%
    \begin{tabular}{lcccccccc}
    \toprule
    Variant &
    A Acc.\,$(\uparrow)$ & B Acc.\,$(\uparrow)$ &
    A Rec.\,$(\uparrow)$ & B Rec.\,$(\uparrow)$ &
    A Prec.\,$(\uparrow)$ & B Prec.\,$(\uparrow)$ &
    A AppRate\,$(\uparrow)$ & B AppRate\,$(\uparrow)$\\
    \midrule
    \multicolumn{9}{c}{\textbf{rep = 0.5 | lbl = 0.0 | A support = 3128 | B support = 2923}}\\
    none                         & \best{0.814} & \second{0.775} & \best{0.939} & \second{0.801} & 0.794 & \second{0.444} & \best{0.724} & 0.338\\
    reweight\_auto                & 0.793 & 0.755 & 0.856 & 0.732 & \best{0.815} & 0.413 & 0.643 & 0.332\\
    reweight\_manual\_A0.5\_B1.5  & \second{0.802} & \best{0.778} & \second{0.886} & \best{0.816} & \second{0.809} & \best{0.449} & \second{0.669} & \second{0.340}\\
    mitigator\_demographic\_parity & 0.495 & 0.667 & 0.354 & 0.394 & 0.663 & 0.252 & 0.327 & 0.294\\
    mitigator\_equalized\_odds     & 0.573 & 0.665 & 0.541 & 0.524 & 0.694 & 0.286 & 0.477 & \best{0.344}\\
    \midrule
    \multicolumn{9}{c}{\textbf{rep = 0.5 | lbl = 0.4 | A support = 3118 | B support = 2986}}\\
    none                         & \best{0.833} & \best{0.784} & \second{0.898} & 0.636 & \best{0.868} & \second{0.181} & \second{0.724} & 0.234\\
    reweight\_auto                & 0.802 & 0.728 & \best{0.941} & \best{0.768} & 0.807 & 0.166 & \best{0.815} & \second{0.307}\\
    reweight\_manual\_A0.5\_B1.5  & \second{0.803} & 0.777 & 0.865 & \second{0.672} & \second{0.855} & \best{0.181} & 0.707 & 0.246\\
    mitigator\_demographic\_parity & 0.418 & \second{0.777} & 0.245 & 0.354 & 0.758 & 0.115 & 0.226 & 0.203\\
    mitigator\_equalized\_odds     & 0.607 & 0.683 & 0.620 & 0.571 & 0.773 & 0.116 & 0.561 & \best{0.327}\\
    \midrule
    \multicolumn{9}{c}{\textbf{rep = 0.5 | lbl = 0.5 | A support = 3167 | B support = 2933}}\\
    none                         & 0.752 & \best{0.779} & \best{0.844} & 0.646 & 0.816 & \best{0.162} & \best{0.741} & 0.238\\
    reweight\_auto                & \second{0.773} & 0.738 & \second{0.843} & \best{0.760} & \second{0.840} & \second{0.155} & \second{0.720} & 0.293\\
    reweight\_manual\_A0.5\_B1.5  & \best{0.777} & \second{0.748} & 0.838 & 0.651 & \best{0.849} & 0.144 & 0.708 & 0.270\\
    mitigator\_demographic\_parity & 0.456 & 0.649 & 0.356 & 0.434 & 0.757 & 0.076 & 0.337 & \second{0.343}\\
    mitigator\_equalized\_odds     & 0.646 & 0.588 & 0.702 & \second{0.669} & 0.783 & 0.092 & 0.643 & \best{0.432}\\
    \midrule
    \multicolumn{9}{c}{\textbf{rep = 0.5 | lbl = 0.6 | A support = 3046 | B support = 3025}}\\
    none                         & 0.785 & \second{0.761} & 0.835 & 0.427 & \second{0.863} & 0.100 & 0.701 & 0.231\\
    reweight\_auto                & \best{0.831} & 0.753 & \second{0.894} & 0.500 & \best{0.875} & \second{0.110} & \second{0.740} & 0.247\\
    reweight\_manual\_A0.5\_B1.5  & \second{0.824} & 0.739 & \best{0.917} & \best{0.555} & 0.851 & \best{0.113} & \best{0.781} & \second{0.267}\\
    mitigator\_demographic\_parity & 0.378 & \best{0.786} & 0.205 & 0.226 & 0.760 & 0.066 & 0.195 & 0.184\\
    mitigator\_equalized\_odds     & 0.633 & 0.574 & 0.695 & \second{0.537} & 0.775 & 0.068 & 0.649 & \best{0.430}\\
    \midrule
    \multicolumn{9}{c}{\textbf{rep = 0.6 | lbl = 0.0 | A support = 3781 | B support = 2346}}\\
    none                         & 0.777 & \second{0.758} & \second{0.873} & 0.734 & 0.775 & 0.412 & \best{0.661} & 0.329\\
    reweight\_auto                & \second{0.796} & 0.756 & 0.859 & \best{0.889} & \best{0.807} & \second{0.423} & 0.625 & \best{0.388}\\
    reweight\_manual\_A0.5\_B1.5  & \best{0.796} & \best{0.776} & \best{0.886} & \second{0.831} & \second{0.792} & \best{0.443} & \second{0.656} & \second{0.347}\\
    mitigator\_demographic\_parity & 0.540 & 0.694 & 0.394 & 0.358 & 0.689 & 0.261 & 0.336 & 0.254\\
    mitigator\_equalized\_odds     & 0.583 & 0.709 & 0.542 & 0.473 & 0.683 & 0.311 & 0.466 & 0.281\\
    \midrule
    \multicolumn{9}{c}{\textbf{rep = 0.6 | lbl = 0.4 | A support = 3818 | B support = 2349}}\\
    none                         & \second{0.811} & \second{0.744} & \second{0.897} & 0.667 & \best{0.847} & 0.182 & 0.757 & 0.281\\
    reweight\_auto                & 0.775 & \best{0.750} & 0.883 & \best{0.761} & 0.817 & \best{0.201} & \second{0.773} & 0.290\\
    reweight\_manual\_A0.5\_B1.5  & \best{0.822} & 0.738 & \best{0.922} & 0.717 & \second{0.844} & \second{0.186} & \best{0.781} & 0.295\\
    mitigator\_demographic\_parity & 0.633 & 0.444 & 0.736 & \second{0.744} & 0.746 & 0.096 & 0.705 & \best{0.593}\\
    mitigator\_equalized\_odds     & 0.660 & 0.573 & 0.749 & 0.700 & 0.769 & 0.117 & 0.697 & \second{0.458}\\
    \midrule
    \multicolumn{9}{c}{\textbf{rep = 0.6 | lbl = 0.5 | A support = 3830 | B support = 2351}}\\
    none                         & 0.824 & \second{0.726} & 0.903 & 0.639 & \best{0.854} & \second{0.115} & 0.742 & 0.289\\
    reweight\_auto                & \second{0.825} & 0.684 & \second{0.914} & \best{0.705} & 0.848 & 0.109 & \second{0.756} & 0.337\\
    reweight\_manual\_A0.5\_B1.5  & \best{0.834} & \best{0.728} & \best{0.929} & \second{0.672} & \second{0.849} & \best{0.120} & \best{0.769} & 0.290\\
    mitigator\_demographic\_parity & 0.537 & 0.531 & 0.548 & 0.574 & 0.725 & 0.062 & 0.531 & \second{0.476}\\
    mitigator\_equalized\_odds     & 0.653 & 0.524 & 0.762 & 0.664 & 0.749 & 0.070 & 0.714 & \best{0.493}\\
    \midrule
    \multicolumn{9}{c}{\textbf{rep = 0.6 | lbl = 0.6 | A support = 3798 | B support = 2319}}\\
    none                         & \second{0.826} & \best{0.737} & \second{0.908} & 0.500 & 0.861 & \best{0.092} & \best{0.771} & 0.263\\
    reweight\_auto                & 0.808 & 0.694 & 0.860 & 0.455 & \best{0.875} & 0.073 & 0.719 & 0.302\\
    reweight\_manual\_A0.5\_B1.5  & \best{0.831} & \second{0.698} & \best{0.910} & 0.509 & \second{0.866} & \second{0.081} & \second{0.768} & 0.303\\
    mitigator\_demographic\_parity & 0.626 & 0.457 & 0.702 & \second{0.714} & 0.767 & 0.061 & 0.669 & \best{0.564}\\
    mitigator\_equalized\_odds     & 0.697 & 0.490 & 0.815 & \best{0.812} & 0.780 & 0.073 & 0.764 & \second{0.540}\\
    \midrule
    \multicolumn{9}{c}{\textbf{rep = 0.7 | lbl = 0.0 | A support = 4374 | B support = 1721}}\\
    none                         & 0.807 & \second{0.749} & \best{0.923} & 0.782 & 0.790 & \second{0.414} & \best{0.697} & 0.358\\
    reweight\_auto                & \best{0.823} & 0.744 & 0.897 & \second{0.794} & \best{0.822} & 0.409 & 0.650 & 0.368\\
    reweight\_manual\_A0.5\_B1.5  & \second{0.812} & \best{0.770} & \second{0.921} & \best{0.801} & \second{0.796} & \best{0.441} & \second{0.690} & 0.344\\
    mitigator\_demographic\_parity & 0.594 & 0.641 & 0.607 & 0.660 & 0.679 & 0.298 & 0.534 & \second{0.419}\\
    mitigator\_equalized\_odds     & 0.637 & 0.622 & 0.752 & 0.715 & 0.676 & 0.295 & 0.664 & \best{0.460}\\
    \midrule
    \multicolumn{9}{c}{\textbf{rep = 0.7 | lbl = 0.4 | A support = 4403 | B support = 1792}}\\
    none                         & \second{0.785} & \second{0.757} & 0.839 & 0.556 & \best{0.853} & \second{0.188} & 0.692 & 0.252\\
    reweight\_auto                & 0.782 & \best{0.810} & \second{0.895} & 0.765 & 0.814 & \best{0.277} & 0.773 & 0.235\\
    reweight\_manual\_A0.5\_B1.5  & \best{0.831} & 0.688 & \best{0.937} & 0.745 & \second{0.841} & 0.180 & \second{0.785} & 0.354\\
    mitigator\_demographic\_parity & 0.654 & 0.472 & 0.761 & \second{0.797} & 0.751 & 0.118 & 0.714 & \second{0.579}\\
    mitigator\_equalized\_odds     & 0.666 & 0.467 & 0.826 & \best{0.810} & 0.734 & 0.118 & \best{0.792} & \best{0.586}\\
    \midrule
    \multicolumn{9}{c}{\textbf{rep = 0.7 | lbl = 0.5 | A support = 4486 | B support = 1745}}\\
    none                         & \second{0.821} & \best{0.720} & 0.903 & 0.546 & \second{0.855} & \second{0.118} & 0.754 & 0.285\\
    reweight\_auto                & 0.819 & 0.653 & \best{0.924} & 0.759 & 0.839 & \best{0.124} & \best{0.786} & 0.379\\
    reweight\_manual\_A0.5\_B1.5  & \best{0.822} & \second{0.682} & 0.895 & 0.583 & \best{0.861} & 0.110 & 0.742 & 0.328\\
    mitigator\_demographic\_parity & 0.675 & 0.359 & 0.821 & \second{0.843} & 0.748 & 0.076 & 0.783 & \best{0.684}\\
    mitigator\_equalized\_odds     & 0.714 & 0.379 & \second{0.904} & \best{0.861} & 0.748 & 0.080 & \second{0.862} & \second{0.665}\\
    \midrule
    \multicolumn{9}{c}{\textbf{rep = 0.7 | lbl = 0.6 | A support = 4553 | B support = 1703}}\\
    none                         & \second{0.847} & \best{0.739} & \second{0.929} & 0.515 & \second{0.866} & 0.112 & 0.766 & 0.263\\
    reweight\_auto                & 0.840 & \second{0.726} & 0.908 & 0.639 & \best{0.873} & \best{0.126} & 0.743 & 0.289\\
    reweight\_manual\_A0.5\_B1.5  & \best{0.848} & 0.713 & \best{0.934} & 0.629 & 0.864 & \second{0.119} & \second{0.772} & 0.301\\
    mitigator\_demographic\_parity & 0.685 & 0.396 & 0.829 & \second{0.753} & 0.754 & 0.068 & \best{0.785} & \best{0.633}\\
    mitigator\_equalized\_odds     & 0.721 & 0.406 & 0.907 & \best{0.856} & 0.753 & 0.077 & \best{0.860} & \second{0.635}\\
    \midrule
    \multicolumn{9}{c}{\textbf{rep = 0.8 | lbl = 0.0 | A support = 5040 | B support = 1186}}\\
    none                         & \second{0.810} & \best{0.777} & 0.917 & 0.767 & \second{0.796} & \best{0.451} & 0.689 & 0.325\\
    reweight\_auto                & \best{0.823} & \second{0.768} & \best{0.942} & \best{0.872} & \best{0.799} & \second{0.446} & \second{0.706} & 0.374\\
    reweight\_manual\_A0.5\_B1.5  & 0.788 & 0.746 & \second{0.922} & \second{0.841} & 0.769 & 0.419 & \best{0.717} & 0.384\\
    mitigator\_demographic\_parity & 0.624 & 0.625 & 0.690 & 0.727 & 0.684 & 0.301 & 0.604 & \second{0.462}\\
    mitigator\_equalized\_odds     & 0.636 & 0.603 & 0.754 & 0.727 & 0.675 & 0.287 & 0.668 & \best{0.484}\\
    \midrule
    \multicolumn{9}{c}{\textbf{rep = 0.8 | lbl = 0.4 | A support = 5100 | B support = 1151}}\\
    none                         & \best{0.845} & \second{0.761} & 0.936 & 0.605 & \second{0.856} & \second{0.177} & 0.767 & 0.255\\
    reweight\_auto                & 0.827 & 0.744 & 0.899 & 0.640 & \best{0.861} & 0.172 & 0.733 & 0.277\\
    reweight\_manual\_A0.5\_B1.5  & \second{0.837} & \best{0.780} & \second{0.945} & 0.663 & 0.842 & \best{0.203} & 0.788 & 0.244\\
    mitigator\_demographic\_parity & 0.716 & 0.411 & 0.890 & \best{0.930} & 0.752 & 0.106 & \second{0.830} & \best{0.653}\\
    mitigator\_equalized\_odds     & 0.731 & 0.424 & \best{0.947} & \second{0.977} & 0.741 & 0.113 & \best{0.896} & \second{0.647}\\
    \midrule
    \multicolumn{9}{c}{\textbf{rep = 0.8 | lbl = 0.5 | A support = 5027 | B support = 1178}}\\
    none                         & \second{0.831} & \second{0.669} & \best{0.946} & 0.673 & 0.839 & \best{0.081} & \second{0.805} & 0.346\\
    reweight\_auto                & 0.821 & 0.650 & 0.902 & 0.510 & \best{0.855} & 0.061 & 0.753 & 0.351\\
    reweight\_manual\_A0.5\_B1.5  & \best{0.835} & \best{0.718} & \second{0.929} & 0.551 & \second{0.853} & \second{0.080} & 0.777 & 0.286\\
    mitigator\_demographic\_parity & 0.702 & 0.441 & 0.849 & \second{0.714} & 0.761 & 0.052 & 0.796 & \second{0.576}\\
    mitigator\_equalized\_odds     & 0.724 & 0.432 & 0.892 & \best{0.776} & 0.762 & 0.055 & \best{0.835} & \best{0.591}\\
    \midrule
    \multicolumn{9}{c}{\textbf{rep = 0.8 | lbl = 0.6 | A support = 5078 | B support = 1112}}\\
    none                         & \second{0.831} & \best{0.729} & 0.922 & 0.527 & \second{0.858} & \best{0.095} & 0.787 & 0.273\\
    reweight\_auto                & \best{0.847} & \second{0.725} & 0.925 & 0.527 & \best{0.874} & \second{0.094} & 0.776 & 0.278\\
    reweight\_manual\_A0.5\_B1.5  & 0.830 & 0.657 & \second{0.935} & 0.618 & 0.849 & 0.086 & 0.807 & 0.354\\
    mitigator\_demographic\_parity & 0.753 & 0.294 & \second{0.949} & \second{0.891} & 0.768 & 0.059 & \second{0.905} & \best{0.745}\\
    mitigator\_equalized\_odds     & 0.752 & 0.301 & \best{0.959} & \best{0.927} & 0.763 & 0.062 & \best{0.920} & \second{0.741}\\
    \bottomrule
    \end{tabular}}%
    \end{table*}   

\end{document}